# 3D pavement surface reconstruction using an RGB-D sensor

A core procedure of pavement management systems is data collection. The modern technologies which are used for this purpose, such as point-based lasers and laser scanners, are too expensive to purchase, operate, and maintain. Thus, it is rarely feasible for city officials in developing countries to conduct data collection using these devices. This paper aims to introduce a cost-effective technology which can be used for pavement distress data collection and 3D pavement surface reconstruction. The applied technology in this research is the Kinect sensor which is not only cost-effective but also sufficiently precise. The Kinect sensor can register both depth and color images simultaneously. A cart is designed to mount an array of Kinect sensors. The cameras are calibrated and the slopes of collected surfaces are corrected via the Singular Value Decomposition (SVD) algorithm. Then, a procedure is proposed for stitching the RGB_D (Red Green Blue–Depth) images using SURF (Speeded-up Robust Features) and MSAC (M-estimator SAmple Consensus) algorithms in order to create a 3D-structure of the pavement surface. Finally, transverse profiles are extracted and some field experiments are conducted to evaluate the reliability of the proposed approach for detecting pavement surface defects.

Ahmadreza Mahmoudzadeh[1], Sayna Firoozi Yeganeh [2], Amir Golroo[3]

[1] Ph.D. Student, Zachry Department of Civil and Environmental Engineering, Texas A&M University, College station, Texas, 77845, US.  A.Mahmoudzadeh@tamu.edu
[1] Ph.D. Student, School of Civil Engineering, College of Engineering, University of Tehran, Tehran, Iran. Sayna.Firoozi@ut.ac.ir
[3] Faculty member, Civil and Environmental Engineering Department, Amirkabir University of Technology, Tehran, Iran. Agolroo@aut.ac.ir

## Introduction

Data collection is an important part of pavement management systems. It is mainly conducted through two methods: manual and semi-automated (or automated). The first method is time-consuming, labor-intensive, and imprecise. The second one is carried out using automated data collection vehicles which are too expensive to implement.

The trade-off between collecting high-quality data and costs of data collection should be considered. There are only a few studies available that used a set of inexpensive and accurate sensors for pavement data collection [1]–[7]. Microsoft Kinect sensors are novel and inexpensive technologies, which can be used for collecting pavement surface distresses data, such as potholes and rutting. Microsoft Kinect V2 is an integrated device containing the RGB and Infrared cameras [8]. This sensor is widely used in different fields of engineering, such as robotic and biomedical engineering, but it is still new in transportation engineering. By considering the capability of this sensor in capturing RGB and depth images simultaneously, it can be used in the field of pavement engineering to generate 3D reconstruction of pavement surfaces in order to collect pavement distress data.

## Methodology

First, a cart was designed and built. The designed cart has the ability to perform data collection in both static and dynamic modes and covers the entire width of a lane (i.e., 3.65 m). By conducting some sample data collection, the collected RGB and depth data were registered. Figure 1 shows a sample of depth data from a rutted pavement surface.

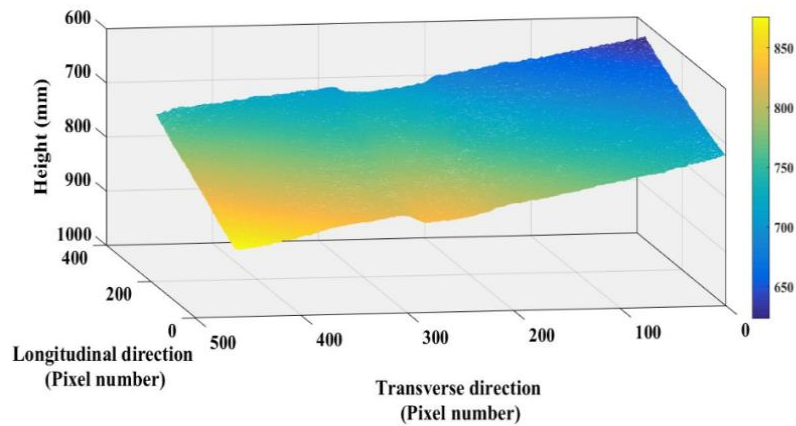

*Figure 1. A sample of depth data from a pavement surface*

The post-processing step contains four major steps; selecting the ROI (region of interest), applying filters for enhancing image quality, correcting the slope, and stitching the depth and RGB images. Firstly, the central part of a depth frame was considered for image processing procedures. After that, a Gaussian filter was applied to the data for reducing the noise effects. Figure 1 shows that the depth image plane is not parallel to a zero-slope pavement surface. So, there is a need to correct the slope. Afterwards, the Singular Value Decomposition (SVD) algorithm was used [9] to correct the slope of collected depth surfaces, as Figure 1 shows that the depth image plane is not parallel to a zero-slope pavement surface.

By calibrating the images and calculating the transformation matrix between the RGB and depth images, the RGB images were stitched. In order to do that, different algorithms such as Speeded-up Robust Features (SURF) [10] and Robust M-estimator SAmple Consensus (MSAC) were used. [11]–[13] Figure 2 to Figure 4 shows how these techniques are working.

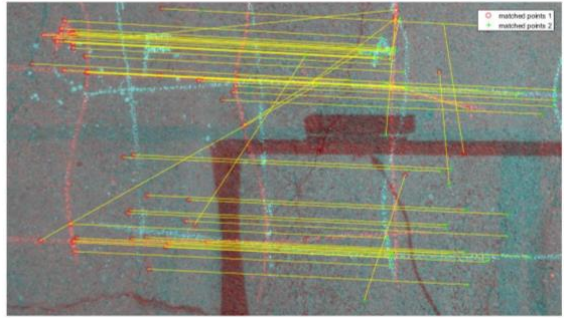

*Figure 2. Finding corresponding features using SURF algorithm*

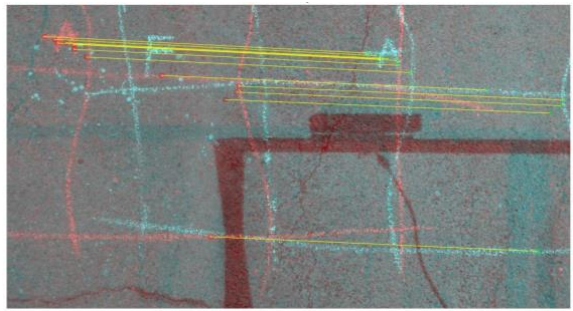

*Figure 3. Remove outliers using MSAC algorithm*

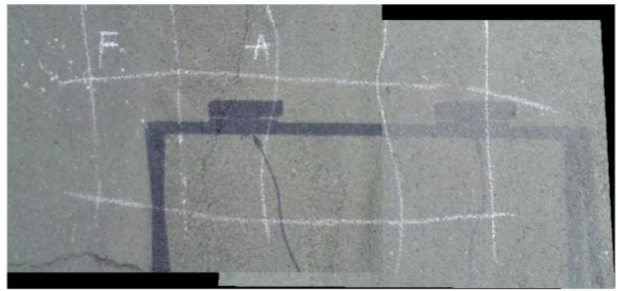

*Figure 4. Stitched RGB images*

## Results and Discussion

By having the stitched RGB images and using the transformation matrix between IR and RGB images, the depth images were stitched. Figure 3 shows the matched (stitched) depth data from a cross section of pavement.

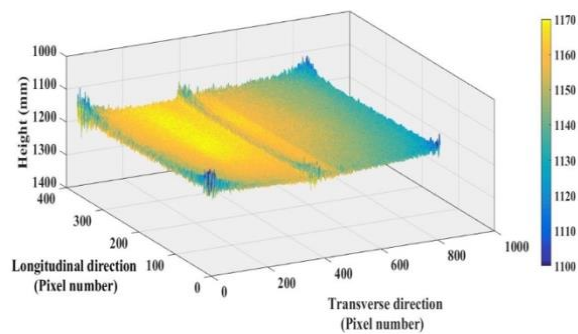

*Figure 5. Stitched depth data*

Finally, the 3D structure of the pavement was constructed. In order to investigate the reliability of the proposed approach, some field experiments were conducted. Figure 6 shows a sample of transverse profiles (extracted from matched depth data) of pavement surface which has rutting distress with different severity levels.

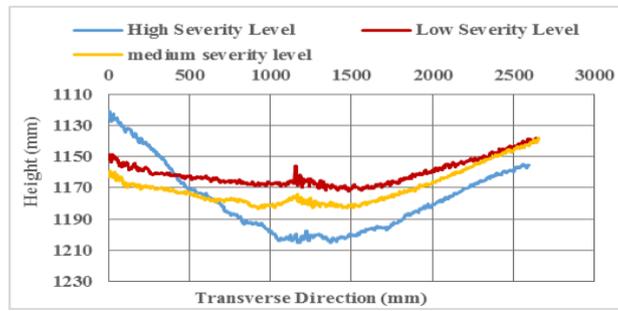
*Figure 6. Transverse profiles of pavement surfaces*

To investigate the accuracy and precision of Kinect sensor, a data collection was performed from asphalt pavement. The distance between the sensor and pavement was measured using an accurate laser distance meter. Figure 7 shows the result of measurement through a Kinect sensor in comparison to the ground truth (laser distance meter measurement). The coefficient of determination shows that the proposed system is accurate enough.

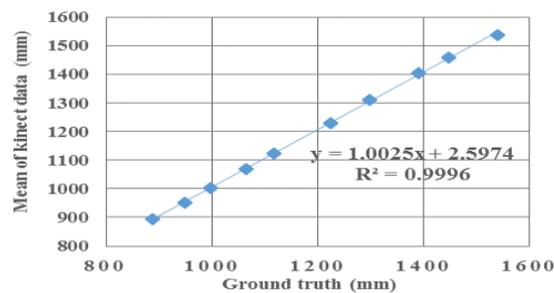
*Figure 7. Estimated distance values versus the ground truth*

Furthermore, to investigate the reliability of the proposed system in collecting the pavement surface defects such as potholes, the dimensions of some artificial defects were measured and compared to the real dimensions. The results show that the mean relative error values are 3.93%, 2.3% and 7.22% for depth, width and length of artificial defects, respectively.

## Conclusions

This paper proposed a cost-effective system which is accurate and precise for collecting pavement distress data and creating a 3D reconstruction of pavement surfaces. The result of this study can be summarized as follows:

- Designing and building an appropriate device on which an array of Kinect sensors can be mounted.
- Stitching the RGB and depth images which were captured from pavement surfaces and generating a 3D structure of pavement.
- Evaluating the reliability of the proposed system.